\documentclass[runningheads]{llncs}
\usepackage{graphicx}
\usepackage{amsmath,amssymb} 
\usepackage{color}

\usepackage{multirow}
\usepackage{booktabs}
\usepackage[pagebackref=false,breaklinks=true,letterpaper=true,colorlinks=true,bookmarks=false,linkcolor=blue,anchorcolor=blue,citecolor=blue]{hyperref}

\def\ie{{\em i.e.}}
\def\eg{{\em e.g.}}
\def\etal{{\em et al. }}

\begin{document}

\title{Occlusion-aware R-CNN: \\ Detecting Pedestrians in a Crowd}
\titlerunning{Occlusion-aware R-CNN: Detecting Pedestrians in a Crowd}

\author{Shifeng Zhang\inst{1,2}\orcidID{0000-0003-3109-5770} \and
Longyin Wen\inst{3}\orcidID{0000-0001-5525-492X} \and\\
Xiao Bian\inst{3}\orcidID{0000-0001-5447-6045} \and
Zhen Lei\inst{1,2}\thanks{Corresponding author}\orcidID{0000-0002-0791-189X} \and\\
Stan Z. Li\inst{4,1,2}\orcidID{0000-0002-2961-8096}}
\authorrunning{S. Zhang, L. Wen, X. Bian, Z. Lei and S. Li}

\institute{Center for Biometrics and Security Research, National Laboratory of Pattern Recognition, Institute of Automation, Chinese Academy of Sciences, Beijing, China\\
\and
University of Chinese Academy of Sciences, Beijing, China\\
\and
GE Global Research, Niskayuna, NY\\
\and
Macau University of Science and Technology, Macau, China\\
\email{\{shifeng.zhang,zlei,szli\}@nlpr.ia.ac.cn, \{longyin.wen,xiao.bian\}@ge.com}
}

\maketitle

\begin{abstract}
Pedestrian detection in crowded scenes is a challenging problem since the pedestrians often gather together and occlude each other. In this paper, we propose a new occlusion-aware R-CNN (OR-CNN) to improve the detection accuracy in the crowd. Specifically, we design a new {\em aggregation loss} to enforce proposals to be close and locate compactly to the corresponding objects. Meanwhile, we use a new part occlusion-aware region of interest (PORoI) pooling unit to replace the RoI pooling layer in order to integrate the prior structure information of human body with visibility prediction into the network to handle occlusion. Our detector is trained in an end-to-end fashion, which achieves state-of-the-art results on three pedestrian detection datasets, \ie, CityPersons, ETH, and INRIA, and performs on-pair with the state-of-the-arts on Caltech.

\keywords{Pedestrian detection \and Occlusion-aware \and Convolutional network \and Structure information \and Visibility prediction}
\end{abstract}

\section{Introduction}

Pedestrian detection is an important research topic in computer vision field with various applications, such as autonomous driving, video surveillance, and robotics, which aims to predict a series of bounding boxes enclosing pedestrian instances in an image. Recent advances in object detection \cite{DBLP:conf/iccv/Girshick15,DBLP:journals/pami/RenHG017,DBLP:conf/nips/DaiLHS16,DBLP:conf/cvpr/LinDGHHB17,DBLP:confs/cvpr/ZhangSF18,DBLP:journals/corr/abs-1712-02408} are driven by the success of deep convolutional neural networks (CNNs), which uses the bounding box regression techniques to accurately localize the objects based on the deep features.

Actually, in real life complex scenarios, occlusion is one of the most significant challenges in detecting pedestrian, especially in the crowded scenes. For example, as pointed out in \cite{DBLP:journals/corr/abs-1711-07752}, $48.8\%$ annotated pedestrians are occluded by other pedestrians in the CityPersons dataset \cite{DBLP:conf/cvpr/ZhangBS17}. Previous methods only require each predicted bounding box to be close to its designated ground truth, without considering the relations among them. Thus, they make the detectors sensitive to the threshold of non-maximum suppression (NMS) in the crowded scenes, wherein filling with occlusions. To that end, Wang \etal \cite{DBLP:journals/corr/abs-1711-07752} design a repulsion loss, which not only pushes each proposal to approach its designated target, but also to keep it away from the other ground truth objects and their corresponding designated proposals. However, it is difficult to control the balance between the repulsion and attraction terms in the loss function to handle the overlapping pedestrians.

In this paper, we propose a new occlusion-aware R-CNN (OR-CNN) based on the Faster R-CNN detection framework \cite{DBLP:journals/pami/RenHG017} to mitigate the impact of occlusion challenge. Specifically, to reduce the false detections of the adjacent overlapping pedestrians, we expect the proposals to be close and locate compactly to the corresponding objects. Thus, inspired by the herd behavior in psychology, we design a new loss function, called {\em aggregation loss} (AggLoss), not only to enforce proposals to be close to the corresponding objects, but also to minimize the internal region distances of proposals associated with the same objects. Meanwhile, to effectively handle partial occlusion, we propose a new part occlusion-aware region of interest (PORoI) pooling unit to replace the original RoI pooling layer in the second stage Fast R-CNN module of the detector, which integrates the prior structure information of human body with visibility prediction into the network. That is, we first partition the pedestrian region into five parts, and pool the features under each part's projection as well as the whole proposal's projection onto the feature map into fixed-length feature vectors by adaptively-sized pooling bins. After that, we use the learned sub-network to predict the visibility score of each part to combine the extracted features for pedestrian detection.

Several experiments are carried out on four pedestrian detection datasets, \ie, CityPersons \cite{DBLP:conf/cvpr/ZhangBS17}, Caltech \cite{DBLP:journals/pami/DollarWSP12}, ETH \cite{DBLP:conf/iccv/EssLG07} and INRIA \cite{DBLP:conf/cvpr/DalalT05}, to demonstrate the superiority of the proposed method, especially for the crowded scenes. Notably, the proposed OR-CNN method achieves the state-of-the-art results with $11.3\%$ MR$^{-2}$ on the CityPersons dataset, $24.5\%$ MR$^{-2}$ on the ETH dataset, and $6.4\%$ MR$^{-2}$ on the INRIA dataset. The main contributions of this work are summarized as follows.
\begin{itemize}
\item We propose a new occlusion-aware R-CNN method, which uses a new designed AggLoss to enforce proposals to be close to the corresponding objects, as well as minimize the internal region distances of proposals associated with the same objects.
\item We design a new PORoI pooling unit to replace the RoI pooling layer in the second Fast R-CNN module to integrate the prior structure information of human body with visibility prediction into the network.
\item Several experiments are carried out on four challenging pedestrian detection datasets, \ie, CityPersons \cite{DBLP:conf/cvpr/ZhangBS17}, Caltech \cite{DBLP:journals/pami/DollarWSP12}, ETH \cite{DBLP:conf/iccv/EssLG07}, and INRIA \cite{DBLP:conf/cvpr/DalalT05}, to demonstrate the superiority of the proposed method.
\end{itemize}

\section{Related Work}

{\flushleft \textbf{Generic Object Detection.}}
Early generic object detectors \cite{DBLP:journals/ijcv/ViolaJ04,DBLP:journals/pami/FelzenszwalbGMR10,DBLP:journals/pami/DollarABP14,DBLP:journals/ijcv/PapageorgiouP00} rely on the sliding window paradigm based on the hand-crafted features and classifiers to find the objects of interest. In recent years, with the advent of deep convolutional neural network (CNN), a new generation of more effective object detection methods based on CNN significantly improve the state-of-the-art performances, which can be roughly divided into two categories, \ie, the one-stage approach and the two-stage approach. The one-stage approach \cite{DBLP:conf/eccv/LiuAESRFB16,DBLP:journals/corr/RedmonF16} directly predicts object class label and regresses object bounding box based on the pre-tiled anchor boxes using deep CNNs. The main advantage of the one-stage approach is its high computational efficiency. In contrast to the one-stage approach, the two-stage approach \cite{DBLP:journals/pami/RenHG017,DBLP:conf/nips/DaiLHS16,DBLP:conf/cvpr/LinDGHHB17} always achieves top accuracy on several benchmarks, which first generates a pool of object proposals by a separated proposal generator (\eg, Selective Search \cite{DBLP:journals/ijcv/UijlingsSGS13}, EdgeBoxes \cite{DBLP:conf/eccv/ZitnickD14}, and RPN \cite{DBLP:journals/pami/RenHG017}), and then predicts the class label and accurate location and size of each proposal.

{\flushleft \textbf{Pedestrian Detection.}}
Even as one of the long-standing problems in computer vision field with an extensive literature, pedestrian detection still receives considerable interests with a wide range of applications. A common paradigm \cite{DBLP:conf/bmvc/DollarTPB09,DBLP:conf/cvpr/YanLYL12,DBLP:conf/cvpr/BenensonMTG13,DBLP:conf/cvpr/YanZLLL13,DBLP:conf/cvpr/ZhangBC14} to address this problem is to train a pedestrian detector that exhaustively operates on the sub-images across all locations and scales. Dalal and Triggs \cite{DBLP:conf/cvpr/DalalT05} design the histograms of oriented gradient (HOG) descriptors and support vector machine (SVM) classifier for human detection. Doll{\'{a}}r \etal \cite{DBLP:journals/pami/DollarABP14} demonstrate that using features from multiple channels can significantly improve the performance. Zhang \etal \cite{DBLP:conf/cvpr/ZhangBS15} provide a systematic analysis for the filtered channel features, and find that with the proper filter bank, filtered channel features can reach top detection quality. Paisitkriangkrai \etal \cite{DBLP:conf/eccv/PaisitkriangkraiSH14} design a new features built on the basis of low-level visual features and spatial pooling, and directly optimize the partial area under the ROC curve for better performance.

Recently, pedestrian detection is dominated by the CNN-based methods (\eg, \cite{DBLP:conf/cvpr/SermanetKCL13,DBLP:conf/cvpr/HosangOBS15,DBLP:conf/cvpr/TianLWT15,DBLP:conf/eccv/CaiFFV16,DBLP:conf/iccv/YangYLL15,DBLP:conf/iccv/BrazilYL17}). Sermanet \etal \cite{DBLP:conf/cvpr/SermanetKCL13} present an unsupervised method using the convolutional sparse coding to pre-train CNN for pedestrian detection. In \cite{DBLP:conf/iccv/CaiSV15}, a complexity-aware cascaded detector is proposed for an optimal trade-off between accuracy and speed. Angelova \etal \cite{DBLP:conf/bmvc/AngelovaKVOF15} combine the ideas of fast cascade and a deep network to detect pedestrian. Yang \etal \cite{DBLP:conf/cvpr/YangCL16} use scale-dependent pooling and layer-wise cascaded rejection classifiers to detect objects efficiently. Zhang \etal \cite{DBLP:conf/eccv/ZhangLLH16} present an effective pipeline for pedestrian detection via using RPN followed by boosted forests. To jointly learn pedestrian detection with the given extra features, a novel network architecture is presented in \cite{DBLP:conf/cvpr/MaoXJC17}. Li \etal \cite{DBLP:journals/tm/li2017scale} use multiple built-in sub-networks to adaptively detect pedestrians across scales. Brazil \etal \cite{DBLP:conf/iccv/BrazilYL17} exploit weakly annotated boxes via a segmentation infusion network to achieve considerable performance gains.

However, occlusion still remains one of the most significant challenges in pedestrian detection, which increases the difficulty in pedestrian localization. Several methods \cite{DBLP:conf/cvpr/OuyangW12,DBLP:conf/iccv/OuyangW13,DBLP:conf/iccv/TianLWT15,DBLP:conf/iccv/MathiasBTG13,DBLP:conf/accv/ZhouY16,DBLP:conf/iccv/WuN05,DBLP:conf/cvpr/ShetNRD07,DBLP:conf/cvpr/EnzweilerESG10,DBLP:conf/eccv/DuanAL10} use part-based model to describe the pedestrian in occlusion handling, which learn a series of part detectors and design some mechanisms to fuse the part detection results to localize partially occluded pedestrians. Besides the part-based model, Leibe \etal \cite{DBLP:conf/cvpr/LeibeSS05} propose an implicit shape model to generate a set of pedestrian hypotheses that are further refined to obtain the visible regions. Wang \etal \cite{DBLP:conf/iccv/WangHY09} divide the template of pedestrian into a set of blocks and conduct occlusion reasoning by estimating the visibility status of each block. Ouyang \etal \cite{DBLP:conf/cvpr/OuyangW13} exploit multi-pedestrian detectors to aid single-pedestrian detectors to handle partial occlusions, especially when the pedestrians gather together and occlude each other in real-world scenarios. In \cite{DBLP:conf/bmvc/TangAS12,DBLP:conf/cvpr/PepikSGS13}, a set of occlusion patterns of pedestrians are discovered to learn a mixture of occlusion-specific detectors. Zhou \etal \cite{DBLP:conf/iccv/ZhouY17} propose to jointly learn part detectors so as to exploit part correlations and reduce the computational cost. Wang \etal \cite{DBLP:journals/corr/abs-1711-07752} introduce a novel bounding box regression loss to detect pedestrians in the crowd scenes. Although numerous pedestrian detection methods are presented in literature, how to robustly detect each individual pedestrian in crowded scenarios is still one of the most critical issues for pedestrian detectors.

\section{Occlusion-aware R-CNN}
Our occlusion-aware R-CNN detector follows the adaptive Faster R-CNN detection framework \cite{DBLP:conf/cvpr/ZhangBS17} for pedestrian detection, with the new designed aggregation loss (Section \ref{sec:aggregation-loss}), and the PORoI pooling unit (Section \ref{sec:poroi-unit}). Specifically, Faster R-CNN \cite{DBLP:journals/pami/RenHG017} consists of two modules, \ie, the first region proposal network (RPN) module and the second Fast R-CNN module. The RPN module is designed to generate high-quality region proposals, and the Fast R-CNN module is used to classify and regress the accurate locations and sizes of objects, based on the generated proposals.

To effectively generate accurate region proposals in the first RPN module, we design the AggLoss term to enforce the proposals locate closely and compactly to the ground-truth object, which is defined as
\begin{equation}
\begin{array}{ll}
\mathbb{L}_{\text{rpn}}(\{ p_i \}, \{ t_i \}, \{ p_i^\ast \}, \{ t_i^\ast \}) =  {\cal L}_{\text{cls}}(\{p_i\}, \{p_i^\ast\}) + \alpha \cdot {\cal L}_{\text{agg}}(\{ p_i^\ast \}, \{ t_i \}, \{ t_i^\ast \}),
\end{array}
\end{equation}
where $i$ is the index of anchor in a mini-batch, $p_i$ and $t_i$ are the predicted confidence of the $i$-th anchor being a pedestrian and the predicted coordinates of the pedestrian, $p_i^\ast$ and $t_i^\ast$ are the associated ground truth class label and coordinates of the $i$-th anchor, $\alpha$ is the hyperparameters used to balance the two loss terms, ${\cal L}_{\text{cls}}(\{p_i\}, \{p_i^\ast\})$ is the classification loss, and ${\cal L}_{\text{agg}}(\{ p_i^\ast \}, \{ t_i \}, \{ t_i^\ast \})$ is the AggLoss (see Section \ref{sec:aggregation-loss}). We use the log loss to calculate the classification loss over two classes (pedestrian $p_i^\ast=1$ {\em vs.} background $p_i^\ast=0$), \ie,
\begin{equation}
\begin{array}{ll}
{\cal L}_{\text{cls}}(\{p_i\},\{p_i^\ast\})=\frac{1}{N_{\text{cls}}}\sum_{i}-\Big(p_i^\ast\log{p_i}+(1-p_i^\ast)\log{(1-p_i)}\Big),
\end{array}
\end{equation}
where $N_{\text{cls}}$ is the total number of anchors in classification.

\subsection{Aggregation Loss}
\label{sec:aggregation-loss}
To reduce the false detections of the adjacent overlapping pedestrians, we enforce proposals to be close and locate compactly to the corresponding ground truth objects. To that end, we design a new aggregation loss (AggLoss) for both the region proposal network (RPN) and Fast R-CNN \cite{DBLP:conf/iccv/Girshick15} modules in the Faster R-CNN algorithm, which is a multi-task loss pushing proposals to be
close to the corresponding ground truth object, while minimizing the internal region distances of proposals associated with the same objects, \ie,
\begin{equation}
\begin{array}{ll}
{\cal L}_{\text{agg}}(\{ p_i^\ast \}, \{\{ t_i \}, \{ t_i^\ast \}) = {\cal L}_{\text{reg}}(\{ p_i^\ast \}, \{ t_i \}, \{ t_i^\ast \}) + \beta \cdot {\cal L}_{\text{com}}(\{ p_i^\ast \}, \{t_i\}, \{t_i^\ast\}),
\end{array}
\label{eq:total-loss}
\end{equation}
where ${\cal L}_{\text{reg}}(\{ p_i^\ast \}, \{ t_i \}, \{ t_i^\ast \})$ is the regression loss which requires each proposal to approach the designated ground truth, and ${\cal L}_{\text{com}}(\{ p_i^\ast \}, \{t_i\}, \{t_i^\ast\})$ is the compactness loss which enforces proposals locate compactly to the designated ground truth object, and $\beta$ is the hyper-parameters used to balance the two loss terms.

Similar to Fast R-CNN \cite{DBLP:conf/iccv/Girshick15}, we use the smooth L1 loss as the regression loss ${\cal L}_{\text{reg}}(\{ p_i^\ast \}, \{t_i\}, \{t_i^\ast\})$ to measure the accuracy of predicted bounding boxes, \ie,
\begin{equation}
\begin{array}{ll}
{\cal L}_{\text{reg}}(\{ p_i^\ast \}, \{ t_i \}, \{ t_i^\ast \})=\frac{1}{N_{\text{reg}}}\sum_{i}p_{i}^\ast\Delta(t_i - t_i^\ast),
\end{array}
\end{equation}
where $N_{\text{reg}}$ is the total number of anchors in regression, and $\Delta(t_i - t_i^\ast)$ is the smooth L1 loss of the predicted bounding box $t_i$.

The compactness term ${\cal L}_{\text{com}}(\{ p_i^\ast \}, \{t_i\}, \{t_i^\ast\})$ is designed to consider the attractiveness among proposals associated with the same ground truth object. In this way, we can make the proposals to locate compactly around the ground truth to reduce the false detections of adjacent overlapping objects. Specifically, we set $\{ {\tilde{t}^\ast}_1, \cdots, \tilde{t}^\ast_{\rho} \}$ to be the ground truth set associated with more than one anchor, and $\{ \Phi_1, \cdots, \Phi_{\rho} \}$ to be the index sets of the associated anchors corresponding to the ground truth objects, \ie, the anchors indexed by $\Phi_k$ are associated to the ground truth $\tilde{t}^\ast_k$, where $\rho$ is the total number of ground-truth object associated with more than one anchor. Thus, we have $\tilde{t}^\ast_k\in\{t_i^\ast\}$, for $k=1,\cdots,\rho$, and $\Phi_i\cap{\Phi_j}=\emptyset$. We use the smooth L1 loss to measure the difference between the average predictions of the anchors indexed by each set in $\{ \Phi_1, \cdots, \Phi_{\rho} \}$ and the corresponding ground truth object, describing the compactness of predicted bounding boxes with respect to the ground truth object, \ie,
\begin{equation}
\begin{array}{ll}
{\cal L}_{\text{com}}(\{ p_i^\ast \}, \{t_i\}, \{t_i^\ast\})=\frac{1}{N_{\text{com}}}\sum_{i=1}^{\rho}\Delta(\tilde{t}_i^\ast - \frac{1}{|\Phi_i|}\sum_{j\in{\Phi_i}}t_j),
\end{array}
\end{equation}
where $N_{\text{com}}$ is the total number of ground truth object associated with more than one anchor (\ie, $N_{\text{com}}=\rho$), and $|\Phi_i|$ is the number of anchors associated with the $i$-th ground truth object.

\begin{figure*}[t]
\centering
\includegraphics[width=0.95\linewidth]{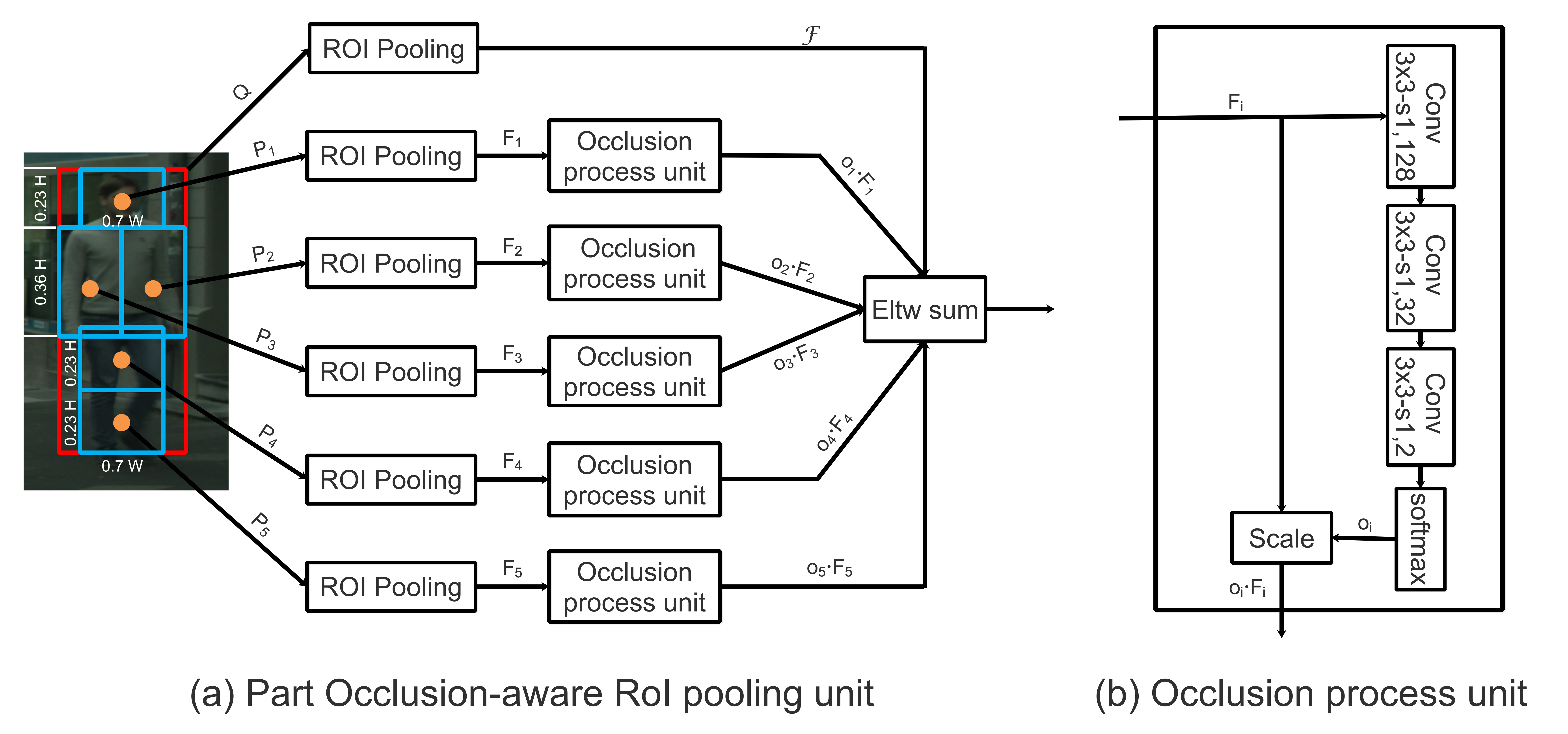}
\caption{For each proposal $Q$, we divide it into $5$ parts ($P_1, \cdots, P_5$) and use RoIPooling to get the features ($F_1, \cdots, F_5$), then feed them into the occlusion process unit to predict the visibility scores ($o_1, \cdots, o_5$). We also apply RoIPooling on $Q$ to generate the holistic feature ${\cal F}$. The final features is computed as ${\cal F} \oplus (o_1 \cdot F_1) \oplus (o_2 \cdot F_2) \oplus (o_3 \cdot F_3) \oplus (o_4 \cdot F_4) \oplus (o_5 \cdot F_5)$ for subsequent classification and regression.}
\label{fig:poroi-unit}
\end{figure*}

\subsection{Part Occlusion-aware RoI Pooling Unit}
\label{sec:poroi-unit}
In real life complex scenarios, occlusion is ubiquitous challenging the accuracy of detectors, especially in crowded scenes. As indicated in \cite{DBLP:conf/cvpr/OuyangW12,DBLP:conf/iccv/TianLWT15,DBLP:conf/iccv/ZhouY17}, the part-based model is effective in handling occluded pedestrians. In contrast to the aforementioned methods, we design a new part occlusion-aware RoI pooling unit to integrate the prior structure information of human body with visibility prediction into the Fast R-CNN module of the detector, which assembles a micro neural network to estimate the part occlusion status. As shown in Figure \ref{fig:poroi-unit} (a), we first divide the pedestrian region into five parts with the empirical ratio in \cite{DBLP:journals/pami/FelzenszwalbGMR10}. For each part, we use the RoI pooling layer \cite{DBLP:conf/iccv/Girshick15} to pool the features into a small feature map with a fixed spatial extent of $H\times{W}$ (\eg, $7\times7$).

We introduce an occlusion process unit, shown in Figure \ref{fig:poroi-unit} (b), to predict the visibility score of the corresponding part based on the pooled features. Specifically, the occlusion process unit is constructed by three convolutional layers followed by a softmax layer with the log loss in training. Symbolically, $c_{i,j}$ indicates the $j$-th part of the $i$-th proposal, $o_{i,j}$ represents its predicted visibility score, and $o_{i,j}^\ast$ is the corresponding ground truth visibility score. If half of the part $c_{i,j}$ is visible, $o_{i,j}^\ast=1$, otherwise $o_{i,j}^\ast=0$. Mathematically, if the intersection between $c_{i,j}$ and the visible region of ground truth object divided by the area of $c_{i,j}$ is larger than the threshold $0.5$, $o_{i,j}^\ast=1$, otherwise $o_{i,j}^\ast=0$. That is
\begin{equation}
o_{i,j}^\ast=\left\{
\begin{aligned}
1&    &\frac{\Omega\big(U(c_{i,j})\cap{V(t_i^\ast)}\big)}{\Omega\big(U(c_{i,j})\big)}>\theta, \\
0&    &\frac{\Omega\big(U(c_{i,j})\cap{V(t_i^\ast)}\big)}{\Omega\big(U(c_{i,j})\big)}\leq\theta,
\end{aligned}
\right.
\end{equation}
where $\Omega(\cdot)$ is the area computing function, $U(c_{i,j})$ is the region of $c_{i,j}$, $V(t_i^\ast)$ is the visible region of the ground truth object $t_i^\ast$, and $\cap$ is the intersection operation between two regions. Then, the loss function of the occlusion process unit is calculated as ${\cal L}_{\text{occ}}(\{ t_i \}, \{ t_i^\ast \})=\sum_{j=1}^{5}-(o_{i,j}^\ast\log{o_{i,j}}+(1-o_{i,j}^\ast)\log({1-o_{i,j}}))$.

After that, we apply the element-wise multiplication operator to multiply the pooled features of each part and the corresponding predicted visibility score to generate the final features with the dimensions $512\times7\times7$. The element-wise summation operation is further used to combine the extracted features of the five parts and the whole proposal for classification and regression in the Fast R-CNN module (see Figure \ref{fig:poroi-unit}).

To further improve the regression accuracy, we also use AggLoss in the Fast R-CNN module, which is defined as:
\begin{equation}
\begin{array}{ll}
\mathbb{L}_{\text{frc}}(\{ p_i \}, \{ t_i \}, \{ p_i^\ast \}, \{ t_i^\ast \}) &=  {\cal L}_{\text{cls}}(\{p_i\}, \{p_i^\ast\}) + \alpha \cdot {\cal L}_{\text{agg}}(\{ p_i^\ast \}, \{ t_i \}, \{ t_i^\ast \}) \\
&+ \lambda \cdot {\cal L}_{\text{occ}}(\{ t_i \}, \{ t_i^\ast \}),
\end{array}
\end{equation}
where $\alpha$ and $\lambda$ are used to balance the three loss terms, ${\cal L}_{\text{cls}}(\{p_i\}, \{p_i^\ast\})$ and ${\cal L}_{\text{agg}}(\{ p_i^\ast \}, \{ t_i \}, \{ t_i^\ast \})$ are the classification and aggregation losses, defined the same as that in the RPN module, and ${\cal L}_{\text{occ}}(\{ t_i \}, \{ t_i^\ast \})$ is the occlusion process loss.

\section{Experiments}
Several experiments are conducted on four datasets: CityPersons \cite{DBLP:conf/cvpr/ZhangBS17}, Caltech-USA \cite{DBLP:journals/pami/DollarWSP12}, ETH \cite{DBLP:conf/iccv/EssLG07}, and INRIA \cite{DBLP:conf/cvpr/DalalT05}, to demonstrate the performance of the proposed OR-CNN method.

\subsection{Experimental Setup}
\label{subsec:imp-details}
Our OR-CNN detector follows the adaptive Faster R-CNN framework \cite{DBLP:conf/cvpr/ZhangBS17} and uses VGG-16 \cite{DBLP:journals/corr/SimonyanZ14a} as the backbone network, pre-trained on the ILSVRC CLS-LOC dataset \cite{DBLP:conf/nips/KrizhevskySH12}. To improve the detection accuracy of pedestrians with small scale, we use the method presented in \cite{DBLP:conf/ijcb/abs-1708-05234,DBLP:conf/ccbr/ZhangZLSWL17} to dense the anchor boxes with the height less than $100$ pixels two times, and use the matching strategy in \cite{DBLP:conf/iccv/abs-1708-05237} to associate the anchors and the ground truth objects.

All the parameters in the newly added convolutional layers are randomly initialized by the ``xavier'' method \cite{DBLP:journals/jmlr/GlorotB10}. We optimize the OR-CNN detector using the Stochastic Gradient Descent (SGD) algorithm with $0.9$ momentum and $0.0005$ weight decay, which is trained on $2$ Titan X GPUs with the mini-batch involving $1$ image per GPU. For the Citypersons dataset, we set the learning rate to $10^{-3}$ for the first $40k$ iterations, and decay it to $10^{-4}$ for another $20k$ iterations. For the Caltech-USA dataset, we train the network for $120k$ iterations with the initial learning rate $10^{-3}$ and decrease it by a factor of $10$ after the first $80k$ iterations. All the hyperparameters $\alpha$, $\beta$ and $\lambda$ are empirically set to $1$.

\begin{table*}[t]
\centering
\caption{Pedestrian detection results on the CityPersons validation set. All models are trained on the training set. The scale indicates the enlarge number of original images in training and testing. $\text{MR}^{-2}$ is used to compare the performance of detectors (lower score indicates better performance). The top three results are highlighted in red, blue and green, respectively. }
\small \setlength{\tabcolsep}{3.5pt}
\begin{tabular}{c|cc|c|c|c|ccc}
\toprule[1.5pt]
\multicolumn{3}{c|}{Method}   &Scale &Backbone &{\em Reasonable} &{\em Heavy} &{\em Partial} &{\em Bare} \\
\hline
\multicolumn{3}{c|}{\multirow{2}{*}{Adapted Faster RCNN \cite{DBLP:conf/cvpr/ZhangBS17}}}                           &$\times1$ &VGG-16 &15.4 &- &- &- \\
\multicolumn{3}{c|}{}                                                                                                                              &$\times1.3$ &VGG-16 &12.8 &- &- &- \\
\multicolumn{3}{c|}{\multirow{2}{*}{Repulsion Loss \cite{DBLP:journals/corr/abs-1711-07752}}}     &$\times1$ &ResNet-50 &13.2 &56.9 &16.8 &7.6 \\
\multicolumn{3}{c|}{}                                                                                                                &$\times1.3$ &ResNet-50 &\color{green}{11.6} &55.3 &\color{green}{14.8} &7.0 \\
\hline
\hline
\multirow{7}{*}{OR-CNN} &AggLoss &PORoI & & & & & &  \\
\cline{2-3}
&  & &$\times1$ &VGG-16 &14.4 &59.4 &18.4 &7.9  \\
&$\surd$ &$\surd$  &$\times1$ &VGG-16 &12.8 &55.7 &15.3 &6.7  \\
&  & &$\times1.3$ &VGG-16 &12.5 &54.5 &16.8 &6.8  \\
&$\surd$ &  &$\times1.3$ &VGG-16 &\color{blue}{11.4} &\color{blue}{52.6} &\color{blue}{13.8} &\color{blue}{6.2} \\
& &$\surd$  &$\times1.3$ &VGG-16 &11.7 &\color{green}{53.0} &\color{green}{14.8} &\color{green}{6.6} \\

&$\surd$ &$\surd$  &$\times1.3$ &VGG-16 &\color{red}{11.0} &\color{red}{51.3} &\color{red}{13.7} &\color{red}{5.9} \\
\bottomrule[1.5pt]
\end{tabular}
\label{tab:cityperson-val}
\end{table*}

\subsection{CityPersons Dataset}
The CityPersons dataset \cite{DBLP:conf/cvpr/ZhangBS17} is built upon the semantic segmentation dataset Cityscapes \cite{DBLP:conf/cvpr/CordtsORREBFRS16} to provide a new dataset of interest for pedestrian detection. It is recorded across $18$ different cities in Germany with $3$ different seasons and various weather conditions. The dataset includes $5,000$ images ($2,975$ for training, $500$ for validation, and $1,525$ for testing) with $\sim35,000$ manually annotated persons plus $\sim13,000$ ignore region annotations. Both the bounding boxes and visible parts of pedestrians are provided and there are approximately $7$ pedestrians in average per image.

Following the evaluation protocol in CityPersons, we train our OR-CNN detector on the training set, and evaluate it on both the validation and the testing sets. The log miss rate averaged over the false positive per image (FPPI) range of $[10^{-2}, 10^0]$ ($\text{MR}^{-2}$) is used to measure the detection performance (lower score indicates better performance). We use the adaptive Faster R-CNN method \cite{DBLP:conf/cvpr/ZhangBS17} trained by ourselves as the baseline detector, which achieves $12.5$ $\text{MR}^{-2}$ on the validation set with $\times1.3$ scale, sightly better than the reported result ($12.8$ $\text{MR}^{-2}$) in \cite{DBLP:conf/cvpr/ZhangBS17}.

\begin{figure*}[t]
\centering
\includegraphics[width=1.0\linewidth]{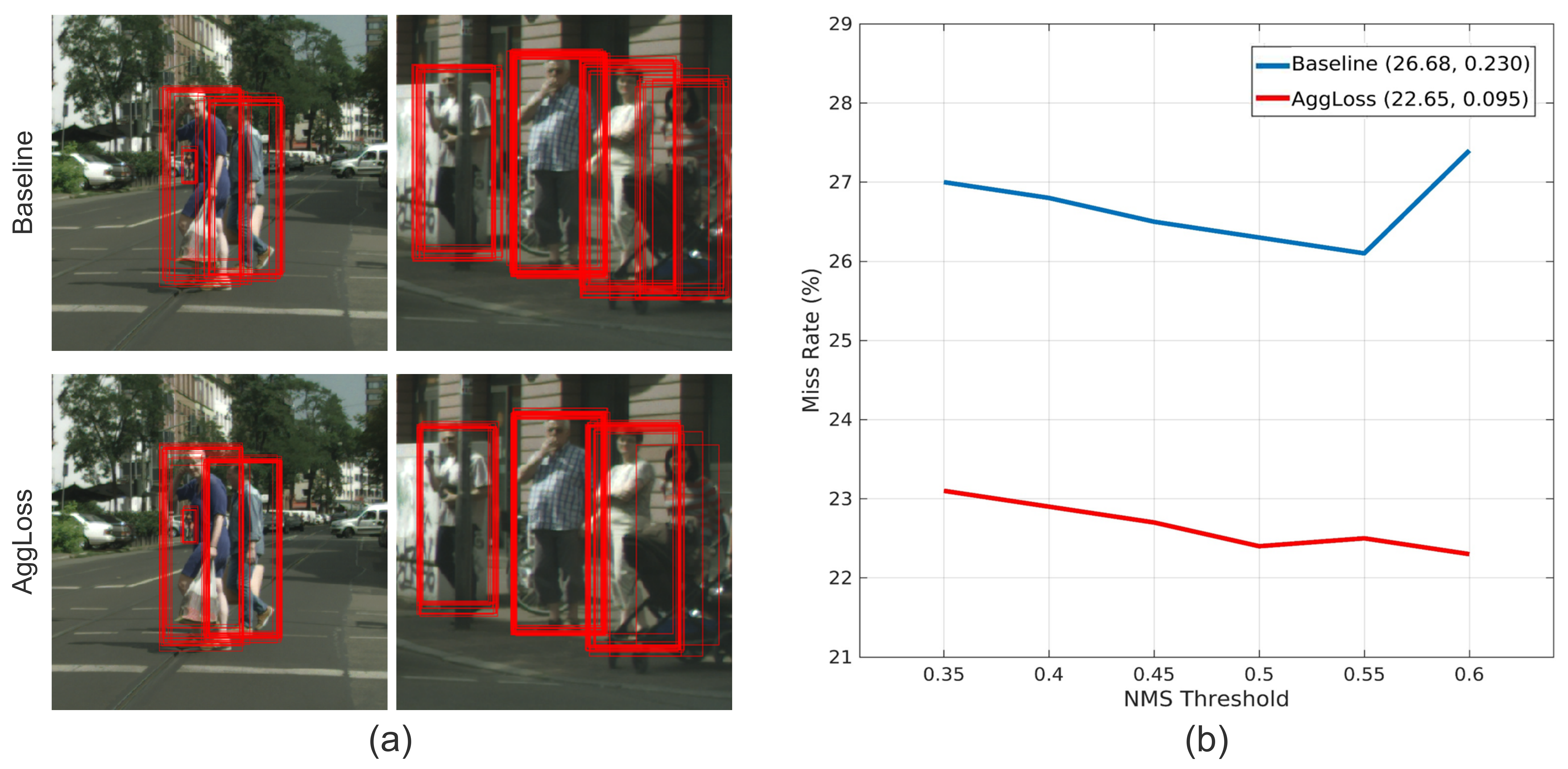}
\caption{(a) Visual comparisons of the predicted bounding boxes before NMS of the baseline and OR-CNN-A detectors. The predictions of OR-CNN-A locate more compactly than that of the baseline detector. (b) Results with AggLoss across various NMS thresholds at $\text{FPPI}=10^{-2}$. The curve of AggLoss is smoother than that of the baseline detector, which indicates that it is less sensitive to the NMS threshold. The scores in the parentheses of the legend are the mean and variance of the miss rate on the curve.}
\label{fig:loss_example}
\end{figure*}

\subsubsection{Ablation Study on AggLoss}

To demonstrate the effectiveness of AggLoss, we construct a detector, denoted as OR-CNN-A, that use AggLoss instead of the original regression loss in the baseline detector \cite{DBLP:conf/cvpr/ZhangBS17}, and evaluate it on the validation set of CityPersons in Table \ref{tab:cityperson-val}. For a fair comparison, we use the same setting of parameters of OR-CNN-A and our OR-CNN detector in both training and testing. All of the experiments are conducted on the reasonable train/validation sets for training and testing.

Comparing the detection results between the baseline and OR-CNN-A in Table \ref{tab:cityperson-val}, we find that using the newly proposed AggLoss can reduce the $\text{MR}^{-2}$ by $1.1\%$ (\ie, $11.4\%$ $\text{MR}^{-2}$ {\em vs.} $12.5\%$ $\text{MR}^{-2}$) with $\times1.3$ scale. It is worth noting that the OR-CNN-A detector achieves $11.4\%$ $\text{MR}^{-2}$ with $\times1.3$ scale, surpassing the state-of-the-art method using Repulsion Loss \cite{DBLP:journals/corr/abs-1711-07752} ($11.6\%$ $\text{MR}^{-2}$), which demonstrates that AggLoss is more effective than Repulsion Loss \cite{DBLP:journals/corr/abs-1711-07752} for detecting the pedestrians in a crowd.

In addition, we also show some visual comparison results of the predicted bounding boxes before NMS of the baseline and OR-CNN-A detectors in Figure \ref{fig:loss_example}(a). As shown in Figure \ref{fig:loss_example}(a), the predictions of OR-CNN-A locate more compactly than that of the baseline detector, and there are fewer predictions of OR-CNN-A lying in between two adjacent ground-truth objects than the baseline detector. This phenomenon demonstrates that AggLoss can push the predictions lying compactly to the ground-truth objects, making the detector less sensitive to the NMS threshold with better performance in the crowd scene. To further validate this point, we also present the results with AggLoss across various NMS threshold at $\text{FPPI}=10^{-2}$ in Figure \ref{fig:loss_example}(b). A high NMS threshold may lead to more false positives, while a low NMS threshold may lead to more false negatives. As shown in Figure \ref{fig:loss_example}(b), we find that the curve of OR-CNN-A is smoother than that of baseline (\ie, the variances of the miss rates are $0.095$ {\em vs.} $0.230$), which indicates that the former is less sensitive to the NMS threshold. It is worth noting that across various NMS thresholds at $\text{FPPI}=10^{-2}$, the OR-CNN-A method always produces lower miss rate, which is due to the NMS operation filtering out more false positives in the predictions of OR-CNN-A than that of baseline, implying that the predicted bounding boxes of OR-CNN-A locate compactly than baseline.

\begin{figure*}[t]
\centering
\includegraphics[width=1.0\linewidth]{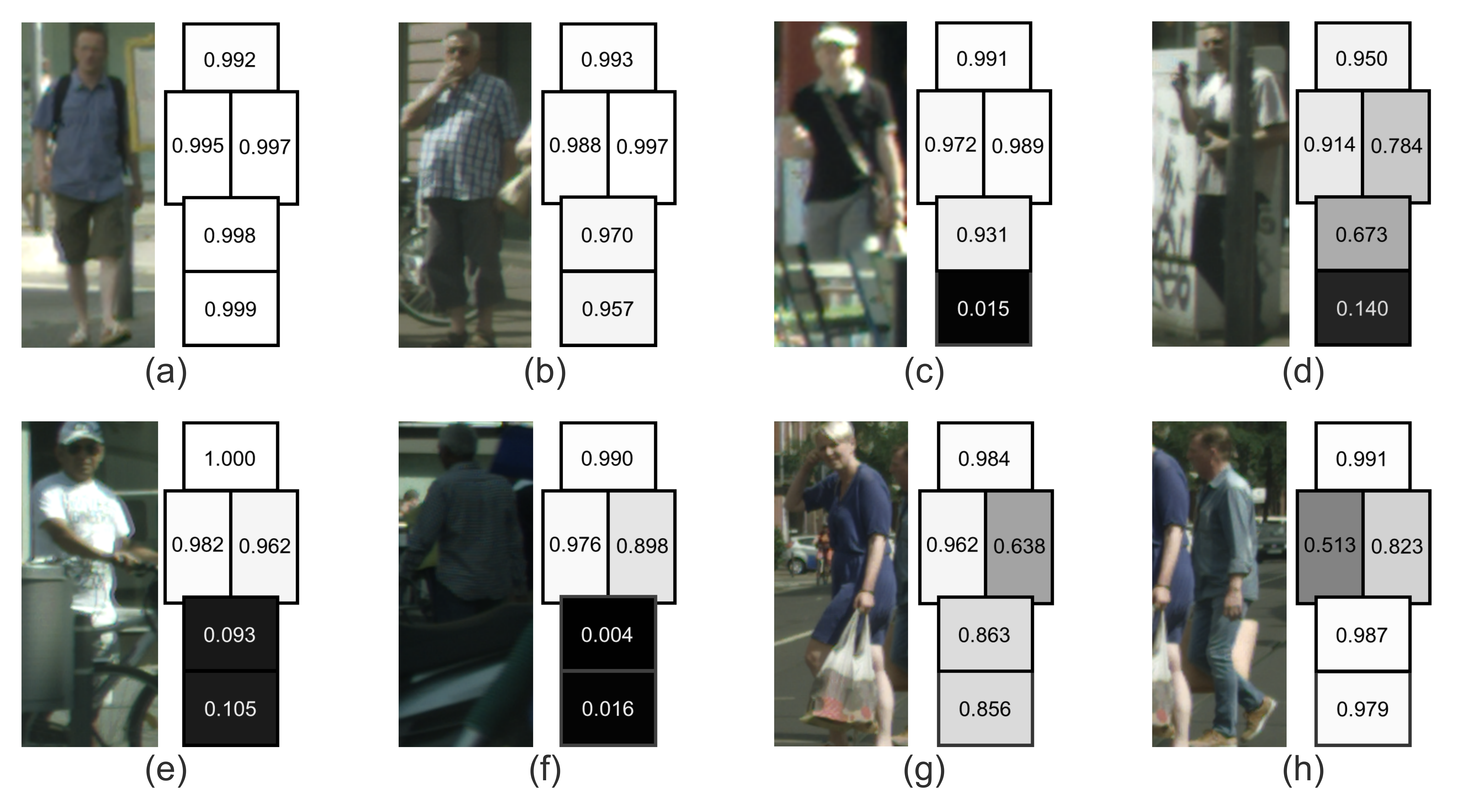}
\caption{Some examples of the predicted visibility scores of the pedestrian parts using the proposed PORoI pooling unit.}
\label{fig:part_example}
\end{figure*}

\subsubsection{Ablation Study on PORoI Pooling}
To validate the effectiveness of the PORoI pooling unit, we construct a detector, denoted as OR-CNN-P, that use the PORoI pooling unit instead of the RoI pooling layer in baseline \cite{DBLP:conf/cvpr/ZhangBS17}, and evaluate it on the validation set of CityPersons in Table \ref{tab:cityperson-val}. For a fair comparison, we use the same parameter settings of OR-CNN-P and our OR-CNN detector in both training and testing. All of the ablation experiments involved CityPersons are conducted on the reasonable train/validation sets for training and testing.

As shown in Table \ref{tab:cityperson-val}, comparing to baseline, OR-CNN-P reduces $0.8\%$ $\text{MR}^{-2}$ with $\times1.3$ scale (\ie, $11.7\%$ {\em vs.} $12.5\%$), which demonstrates the effectiveness of the PORoI pooling unit in pedestrian detection. Meanwhile, we also present some qualitative results of the predictions with the visibility scores of the corresponding parts in Figure \ref{fig:part_example}. Notably, we find that the visibility scores predicted by the PORoI pooling unit are in accordance with the human visual system. As shown in Figure \ref{fig:part_example}(a) and (b), if the pedestrian is not occluded, the visibility score of each part of the pedestrian approaches $1$. However, if some parts of the pedestrians are occluded by the background obstacles or other pedestrians, the scores of the corresponding parts decrease, such as the occluded thigh and calf in Figure \ref{fig:part_example}(c)-(f). Besides, if two pedestrians gather together and occlude each other, our PORoI pooling unit successfully detects the occluded human parts that can help lower the contributions of the occluded parts in pedestrian detection, see Figure \ref{fig:part_example}(g) and (h). Notably, the detection accuracy of the OR-CNN detector can not be improved if we fix the visibility score of each part to $1$ instead of using the predictions of the occlusion process unit (see Figure \ref{fig:poroi-unit}). Thus, the occlusion process unit is the key component to detection accuracy, since it enables our PORoI pooling unit to detect the occluded parts of pedestrians, which is useful to help extract effective features for detection.

\subsubsection{Evaluation Results}
We compare the proposed OR-CNN method\footnote{Due to the shortage of computational resources and the memory issue, we only train OR-CNN with two kinds of input sizes, \ie, $\times1$ and $\times1.3$ scale. We believe the accuracy of OR-CNN can be further improved using larger input images. Thus, we only compare the proposed method with the state-of-the-art detectors using $\times1$ and $\times1.3$ input scales.} with the state-of-the-art detectors \cite{DBLP:journals/corr/abs-1711-07752,DBLP:conf/cvpr/ZhangBS17} on both the validation and testing sets of CityPersons in Table \ref{tab:cityperson-val} and Table \ref{tab:cityperson-test}, respectively. Our OR-CNN achieves the state-of-the-art results on the validation set of CityPersons by reducing $0.6\%$ $\text{MR}^{-2}$ (\ie, $11.0\%$ {\em vs.} $11.6\%$ of \cite{DBLP:journals/corr/abs-1711-07752}) with $\times1.3$ scale and $0.4\%$ $\text{MR}^{-2}$ (\ie, $12.8\%$ {\em vs.} $13.2\%$ of \cite{DBLP:journals/corr/abs-1711-07752}) with $\times1$ scale, surpassing all published approaches \cite{DBLP:journals/corr/abs-1711-07752,DBLP:conf/cvpr/ZhangBS17}, which demonstrates the superiority of the proposed method in pedestrian detection.

\begin{table*}[t]
\centering
\caption{Pedestrian detection results of the proposed OR-CNN method and other state-of-the-art methods on the CityPersons testing set. The scale indicates the enlarge number of original images in training and testing. $\text{MR}^{-2}$ is used to compare of the performance of detectors (lower score indicates better performance).}
\small \setlength{\tabcolsep}{5.5pt}
\begin{tabular}{c|c|c|cc}
\toprule[1.5pt]
Method &Backbone &Scale &{\em Reasonable} &{\em Reasonable-Small} \\
\hline
Adapted FasterRCNN \cite{DBLP:conf/cvpr/ZhangBS17} & VGG-16 &$\times$1.3 &12.97 &37.24 \\
Repulsion Loss \cite{DBLP:journals/corr/abs-1711-07752} & ResNet-50 &$\times$1.5 &11.48 &15.67 \\
\hline
OR-CNN & VGG-16 &\textbf{$\times$1.3} &\textbf{11.32} &\textbf{14.19} \\
\bottomrule[1.5pt]
\end{tabular}
\label{tab:cityperson-test}
\end{table*}

To demonstrate the effectiveness of OR-CNN under various occlusion levels, we follow the strategy in \cite{DBLP:journals/corr/abs-1711-07752} to divide the {\em Reasonable} subset in the validation set (occlusion $<35\%$) into the {\em Reasonable-Partial} subset ($10\%<$ occlusion $\le35\%$), denoted as {\em Partial} subset, and the {\em Reasonable-Bare} subset (occlusion $\le10\%$), denoted as {\em Bare} subset. Meanwhile, we denote the annotated pedestrians with the occlusion ratio larger than $35\%$ (that are not included in the {\em Reasonable} set) as {\em Heavy} subset. We report the results of the proposed OR-CNN method and other state-of-the-art methods \cite{DBLP:journals/corr/abs-1711-07752,DBLP:conf/cvpr/ZhangBS17} on these three subsets in Table \ref{tab:cityperson-val}. As shown in Table \ref{tab:cityperson-val}, OR-CNN outperforms the state-of-the-art methods consistently across all three subsets, \ie, reduces $1.1\%$ $\text{MR}^{-2}$ on the {\em Bare} subset, $1.1\%$ $\text{MR}^{-2}$ on the {\em Partial} subset, and $4.0\%$ $\text{MR}^{-2}$ on the {\em Heavy} subset. Notably, when the occlusion becomes severely (\ie, from {\em Bare} subset to {\em Heavy} subset), the performance improvement of our OR-CNN is more obvious compared to the state-of-the-art methods \cite{DBLP:journals/corr/abs-1711-07752,DBLP:conf/cvpr/ZhangBS17}, which demonstrates that the AggLoss and PORoI pooling unit are extremely effective to address the occlusion challenge.

In addition, we also evaluate the proposed OR-CNN method on the testing set of CityPersons \cite{DBLP:conf/cvpr/ZhangBS17}. Following its evaluation protocol, we submit the detection results of OR-CNN to the authors for evaluation and report the results in Table \ref{tab:cityperson-test}. The proposed OR-CNN method achieves the top accuracy with only $\times1.3$ scale. Although the second best detector Repulsion Loss \cite{DBLP:journals/corr/abs-1711-07752} uses much bigger input images (\ie, $\times1.5$ scale of \cite{DBLP:journals/corr/abs-1711-07752} {\em vs.} $\times1.3$ scale of OR-CNN) and stronger backbone network (\ie, ResNet-50 of \cite{DBLP:journals/corr/abs-1711-07752} {\em vs.} VGG-16 of OR-CNN), it still produces $0.16\%$ higher $\text{MR}^{-2}$ on the {\em Reasonable} subset and $1.48\%$ higher $\text{MR}^{-2}$ on the {\em Reasonable-Small} subset. We believe the performance of  OR-CNN can be further improved by using bigger input images and stronger backbone network.

\begin{figure*}[t]
\centering
\includegraphics[width=0.9\linewidth]{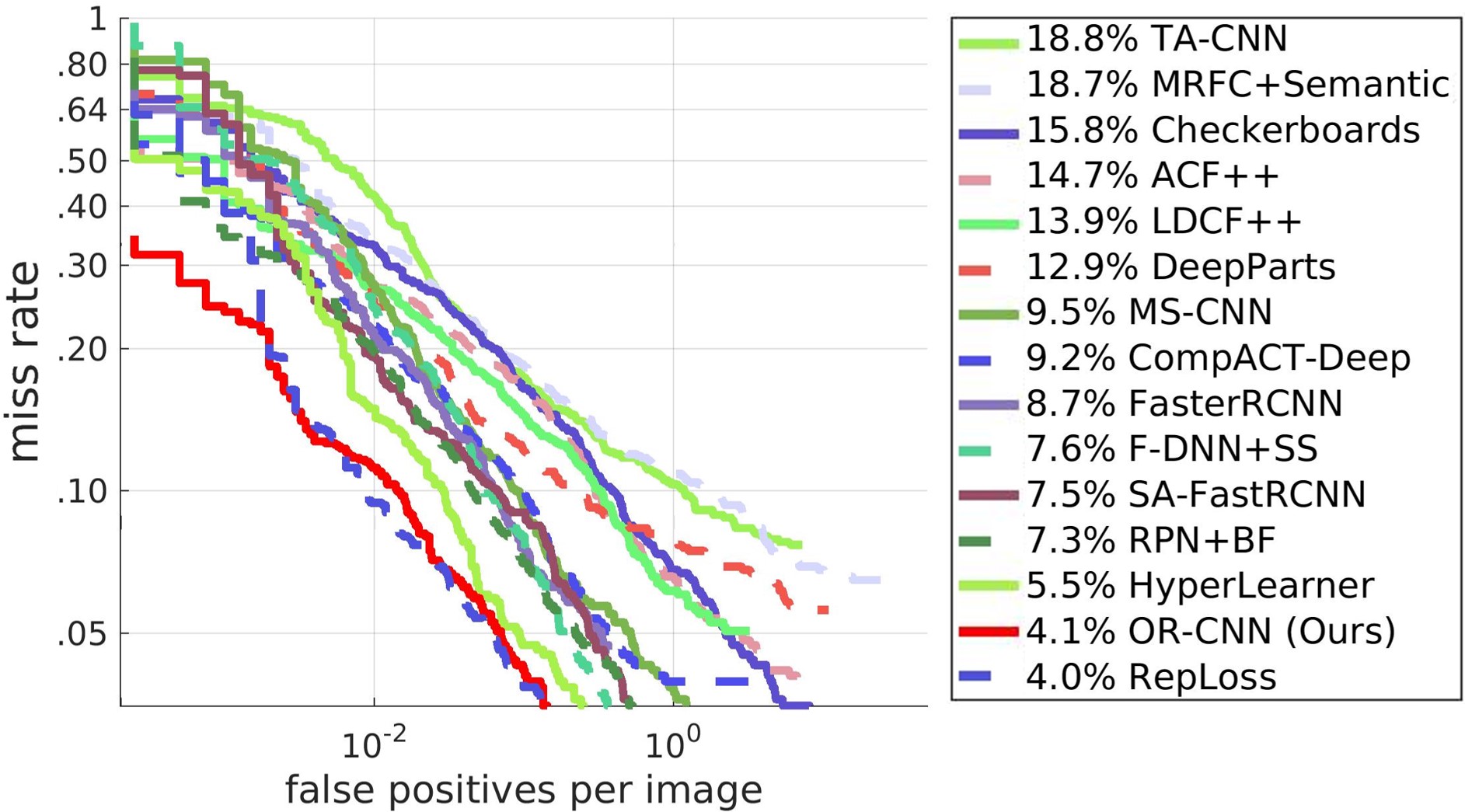}
\caption{Comparisons with the state-of-the-art methods on the Caltech-USA dataset. The scores in the legend are the ${\text{MR}}^{-2}$ scores of the corresponding methods.}
\label{fig:caltech}
\end{figure*}

\subsection{Caltech-USA Dataset}
The Caltech-USA dataset \cite{DBLP:journals/pami/DollarWSP12} is one of the most popular and challenging datasets for pedestrian detection, which comes from approximately $10$ hours $30$Hz VGA video recorded by a car traversing the streets in the greater Los Angeles metropolitan area. We use the new high quality annotations provided by \cite{DBLP:conf/cvpr/ZhangBOHS16} to evaluate the proposed OR-CNN method. The training and testing sets contains $42,782$ and $4,024$ frames, respectively. Following \cite{DBLP:journals/pami/DollarWSP12}, the log-average miss rate over $9$ points ranging from $10^{-2}$ to $10^{0}$ FPPI is used to evaluate the performance of the detectors.

We directly fine-tune the detection models pre-trained on CityPersons \cite{DBLP:conf/cvpr/ZhangBS17} of the proposed OR-CNN method on the training set in Caltech-USA. Similar to \cite{DBLP:journals/corr/abs-1711-07752}, we evaluate the OR-CNN method on the {\em Reasonable} subset of the Caltech-USA dataset, and compare it to other state-of-the-art methods (\eg, \cite{DBLP:journals/corr/abs-1711-07752,DBLP:conf/cvpr/ZhangBS15,DBLP:conf/cvpr/TianLWT15,DBLP:conf/eccv/CaiFFV16,DBLP:conf/iccv/CaiSV15,DBLP:conf/eccv/ZhangLLH16,DBLP:conf/cvpr/MaoXJC17,DBLP:journals/tm/li2017scale,DBLP:conf/iccv/TianLWT15,DBLP:conf/cvpr/CosteaN16,DBLP:conf/icpr/Ohn-BarT16a,DBLP:conf/wacv/DuELD17}) in Figure \ref{fig:caltech}. Notably, the {\em Reasonable} subset (occlusion $<35\%$) only includes the pedestrians with at least $50$ pixels tall, which is widely used to evaluate the pedestrian detectors. As shown in Figure \ref{fig:caltech}, the OR-CNN method performs competitively with the state-of-the-art method \cite{DBLP:journals/corr/abs-1711-07752} by producing $4.1\%$ $\text{MR}^{-2}$.

\subsection{ETH Dataset}
\begin{figure}[t] \centering
\includegraphics[width=0.9\linewidth]{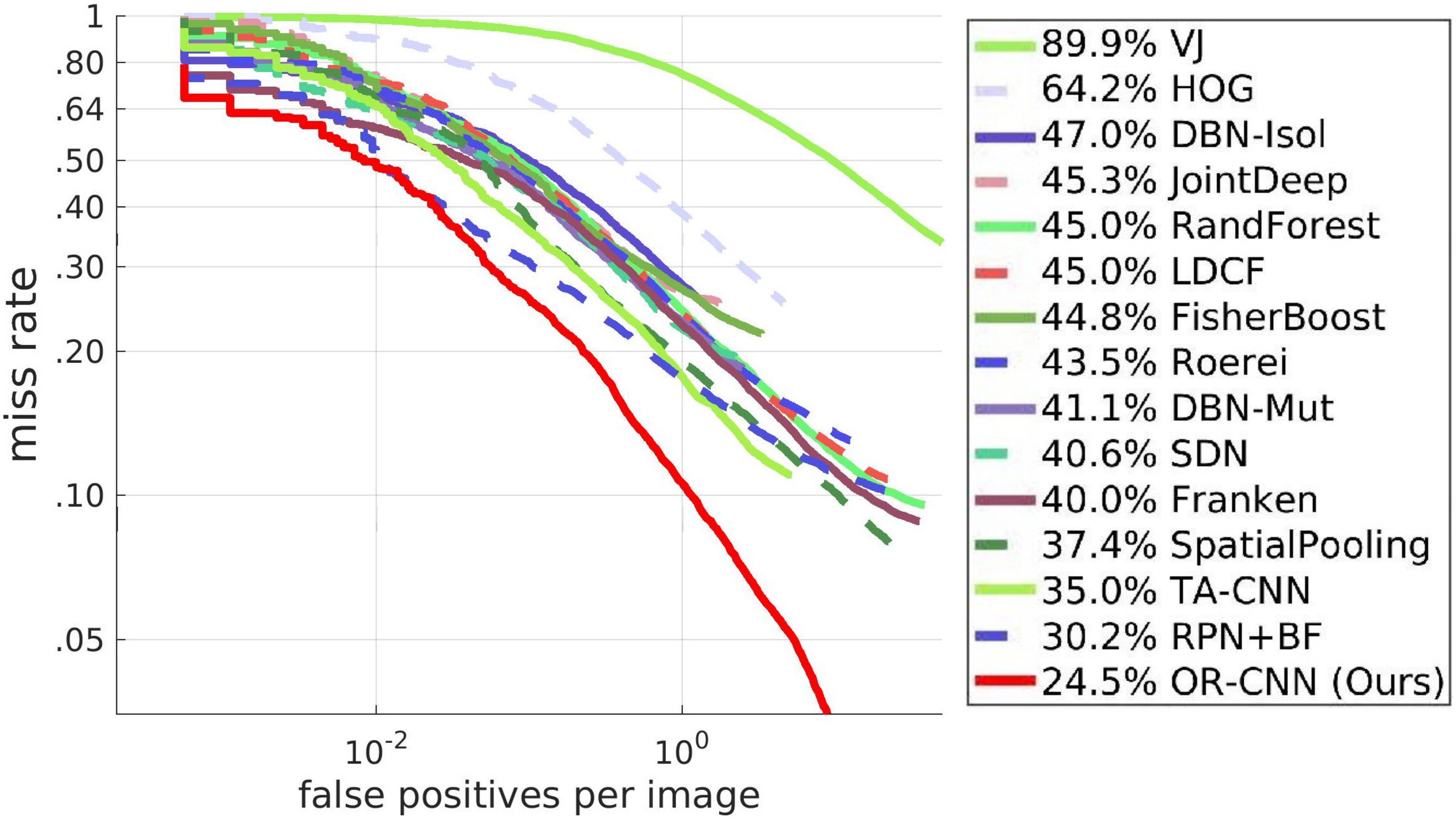}
\caption{Comparisons with the state-of-the-art methods on the ETH dataset. The scores in the legend are the ${\text{MR}}^{-2}$ scores of the corresponding methods.}
\label{fig:eth}
\end{figure}

To verify the generalization capacity of the proposed OR-CNN detector, we directly use the model trained on the CityPersons \cite{DBLP:conf/cvpr/ZhangBS17} dataset to detect the pedestrians in the ETH dataset \cite{DBLP:conf/iccv/EssLG07} without fine-tuning. That is, all $1,804$ frames in three video clips of the ETH dataset \cite{DBLP:conf/iccv/EssLG07} are used to evaluate the performance of the OR-CNN detector. We use $\text{MR}^{-2}$ to evaluate the performance of the detectors, and compare the proposed OR-CNN method with other state-of-the-art methods (\ie, \cite{DBLP:conf/cvpr/DalalT05,DBLP:journals/ijcv/ViolaJ04,DBLP:conf/cvpr/BenensonMTG13,DBLP:conf/eccv/PaisitkriangkraiSH14,DBLP:conf/cvpr/TianLWT15,DBLP:conf/eccv/ZhangLLH16,DBLP:conf/cvpr/OuyangW12,DBLP:conf/iccv/OuyangW13,DBLP:conf/iccv/MathiasBTG13,DBLP:journals/ijcv/ShenWPH13,DBLP:conf/cvpr/OuyangZW13,DBLP:conf/iccv/MarinVLAL13,DBLP:conf/nips/NamDH14,DBLP:conf/cvpr/LuoTWT14}) in Figure \ref{fig:eth}. Our OR-CNN detector achieves the top accuracy by reducing $5.7\%$ $\text{MR}^{-2}$ comparing to the state-of-the-art results (\ie, $24.5\%$ of OR-CNN {\em vs.} $30.2\%$ RFN-BF \cite{DBLP:conf/eccv/ZhangLLH16}). The results on the ETH dataset not only demonstrates the superiority of the proposed OR-CNN method in pedestrian detection, but also verifies its generalization capacity to other scenarios.

\begin{figure}[t]
\centering
\includegraphics[width=0.9\linewidth]{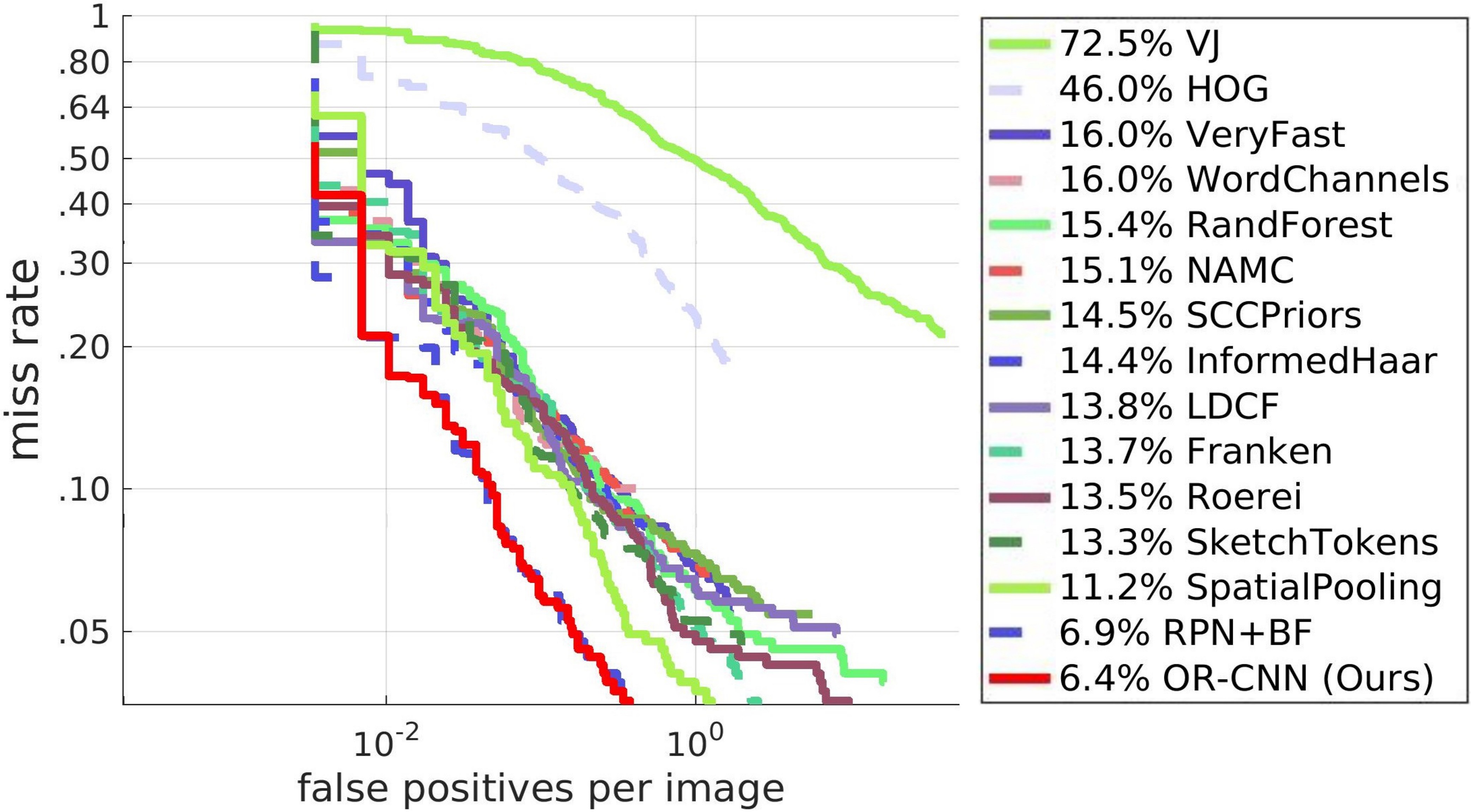}
\caption{Comparisons with the state-of-the-art methods on the INRIA dataset. The scores in the legend are the ${\text{MR}}^{-2}$ scores of the corresponding methods.}
\label{fig:inria}
\end{figure}

\subsection{INRIA Dataset}
The INRIA dataset \cite{DBLP:conf/cvpr/DalalT05} contains images of high resolution pedestrians collected mostly from holiday photos, which consists of $2,120$ images, including $1,832$ images for training and $288$ images. Specifically, there are $614$ positive images and $1,218$ negative images in the training set. We use the $614$ positive images in the training set to fine-tune our model pre-trained on CityPersons for $5k$ iterations, and test it on the $288$ testing images. Figure \ref{fig:inria} shows that our OR-CNN method achieves an $\text{MR}^{-2}$ of $6.4\%$, better than the other available competitors (\ie, \cite{DBLP:conf/cvpr/DalalT05,DBLP:journals/ijcv/ViolaJ04,DBLP:conf/cvpr/BenensonMTG13,DBLP:conf/cvpr/ZhangBC14,DBLP:conf/eccv/PaisitkriangkraiSH14,DBLP:conf/eccv/ZhangLLH16,DBLP:conf/iccv/MathiasBTG13,DBLP:conf/iccv/MarinVLAL13,DBLP:conf/nips/NamDH14,DBLP:conf/cvpr/LimZD13,DBLP:conf/bmvc/YangWW15,DBLP:conf/bmvc/TocaCP15,DBLP:conf/cvpr/CosteaN14,DBLP:conf/cvpr/BenensonMTG12}), which demonstrates the effectiveness of the proposed method in pedestrian detection.

\section{Conclusions}
In this paper, we present a new occlusion-aware R-CNN method to improve the pedestrian detection accuracy in crowded scenes. Specifically, we design a new aggregation loss to reduce the false detections of the adjacent overlapping pedestrians, by simultaneously enforcing the proposals to be close to the associated objects, and locate compactly. Meanwhile, to effectively handle partial occlusion, we propose a new part occlusion-aware RoI pooling unit to replace the RoI pooling layer in the Fast R-CNN module of the detector, which integrates the prior structure information of human body with visibility prediction into the network to handle occlusion. Our method is trained in an end-to-end fashion and achieves the state-of-the-art accuracy on three pedestrian detection datasets, \ie, CityPersons, ETH, and INRIA, and performs on-pair with the state-of-the-arts on Caltech. In the future, we plan to improve the method in two aspects. First, we would like to redesign the PORoI pooling unit to jointly estimate the location, size, and occlusion status of the object parts in the network, instead of using the empirical ratio. And then, we plan to extend the proposed method to detect other kinds of objects, \eg, car, bicycle, tricycle, etc.

\section*{Acknowledgments}
This work was supported by the National Key Research and Development Plan (Grant No.2016YFC0801002), the Chinese National Natural Science Foundation Projects $\#61473291$, $\#61572501$, $\#61502491$, $\#61572536$, the Science and Technology Development Fund of Macau (No. 0025/2018/A1, 151/2017/A, 152/2017/A), JDGrapevine Plan and AuthenMetric R\&D Funds. We also thank NVIDIA for GPU donations through their academic program.

\bibliographystyle{splncs04}
\bibliography{reference}

\end{document}